\documentclass[conference]{IEEEtran}

\makeatletter

\def\ps@IEEEtitlepagestyle{%
  \def\@oddfoot{\mycopyrightnotice}%
  \def\@evenfoot{}%
}
\def\mycopyrightnotice{%
  {\footnotesize XXX-X-XXXX-XXXX-X/XX/\$XX.00~\copyright~20XX IEEE\hfill}
  \gdef\mycopyrightnotice{}
}

\usepackage{blindtext}
\usepackage{multirow}
\usepackage[nocompress]{cite}

\usepackage{eso-pic}
\usepackage{subcaption}
\IEEEoverridecommandlockouts
\usepackage{cite}
\usepackage{amsmath,amssymb,amsfonts}
\usepackage{algorithmic}
\usepackage{graphicx}
\usepackage{algorithm2e}
\usepackage{textcomp}
\usepackage{xcolor}
\def\BibTeX{{\rm B\kern-.05em{\sc i\kern-.025em b}\kern-.08em
    T\kern-.1667em\lower.7ex\hbox{E}\kern-.125emX}}
    
\usepackage{eso-pic}
\newcommand\AtPageUpperMyright[1]{\AtPageUpperLeft{%
 \put(\LenToUnit{0.17\paperwidth},\LenToUnit{-2cm}){%
     \parbox{0.9\textwidth}{\raggedleft\fontsize{8}{11}\selectfont #1}}%
 }}%
\newcommand{\conf}[1]{%
\AddToShipoutPictureBG*{%
\AtPageUpperMyright{#1}
}
}

\begin{document}
\title{\vspace*{1cm}  Anomaly Detection  in Power Generation Plants with Generative Adversarial Networks\\
\thanks{}
}

\author{\IEEEauthorblockN{Marcellin Atemkeng}
\IEEEauthorblockA{\textit{Department of Mathematics, Rhodes University}\\
Grahamstown 6139, South Africa.\\ m.atemkeng@ru.ac.za}
\and
\IEEEauthorblockN{Toheeb A. Jimoh}
\IEEEauthorblockA{\textit{Department of Computer Science and Information Systems} \\
University of Limerick, Limerick, Ireland\\ toheeb.jimoh@ul.ie}
\textit{AIMS, Kigali, Rwanda, jimoh.toheeb@aims.ac.rw}}

\maketitle
\conf{\textit{  III. International Conference on Electrical, Computer and Energy Technologies (ICECET 2023) \\ 
16-17 November 2023, Cape Town-South Africa}}
\begin{abstract}
Anomaly detection is a critical task that involves the identification of data points that deviate from a predefined pattern, useful for fraud detection and related activities. Various techniques are employed for anomaly detection, but recent research indicates that deep learning methods, with their ability to discern intricate data patterns, are well-suited for this task. This study explores the use of Generative Adversarial Networks (GANs) for anomaly detection in power generation plants. 
The dataset used in this investigation comprises fuel consumption records obtained from power generation plants operated by a telecommunications company. The data was initially collected in response to observed irregularities in the fuel consumption patterns of the generating sets situated at the company's base stations. The dataset was divided into anomalous and normal data points based on specific variables, with 64.88\% classified as normal and 35.12\% as anomalous. An analysis of feature importance, employing the random forest classifier, revealed that "Running Time Per Day" exhibited the highest relative importance. A GANs model was trained and fine-tuned both with and without data augmentation, with the goal of increasing the dataset size to enhance performance. The generator model consisted of five dense layers using the $tanh$ activation function, while the discriminator comprised six dense layers, each integrated with a dropout layer to prevent overfitting. Following data augmentation, the model achieved an accuracy rate of 98.99\%, compared to 66.45\% before augmentation. This demonstrates that the model nearly perfectly classified data points into normal and anomalous categories, with the augmented data significantly enhancing the GANs' performance in anomaly detection. 
As a result, this study recommends the use of GANs, particularly when coupled with large datasets, for effective anomaly detection tasks.
\end{abstract}



\section{Introduction}
The telecommunications industry stands as a prominent sector within the Information Communication Technologies (ICT) landscape, heavily reliant on a substantial supply of electrical power for its seamless operations. This dependency is particularly critical, as ICT companies consume approximately 3\% of the world's total electrical energy \cite{humar2011rethinking}. However, the accessibility of reliable electrical power remains a persistent concern, especially in underdeveloped regions, notably across Africa. With the proliferation of telecommunication infrastructure, including base stations across the continent, the industry has been compelled to explore alternative energy sources. These alternatives encompass the use of gasoline or diesel generators and harnessing solar power, among others, to ensure continuous and robust communication networks. 

TeleInfra, a telecommunications company operating in Cameroon, is among those grappling with these challenges due to the nation's erratic power supply. The telecommunication equipment dispersed across both rural and urban areas necessitates a continuous electricity supply to fulfil its mission of establishing resilient communication networks. Unfortunately, the country primarily relies on hydropower (constituting 73\% of its energy generation), which is prone to disruptions, especially during dry seasons marked by low water levels \cite{erasmus}. Moreover, access to electricity is limited, with only approximately 14\% coverage in rural areas and varying from 65\% to 88\% in urban centres. Furthermore, the shift towards alternative power sources, notably the use of generators, has introduced new challenges in the form of irregularities or anomalies in fuel consumption at these base stations. Past research in \cite{ayang}, suggests that various factors such as the mismanagement of air-conditioning and lighting systems, as well as building design, may contribute to the heightened power consumption at these sites.

Anomalies are defined as data points that deviate from the expected pattern within a dataset, often representing a distinct distribution within the broader dataset. It's worth noting that there are varying definitions and misconceptions about anomalies, which have been comprehensively addressed in \cite{anomaly_review}. This work offers a comprehensive overview of different typologies and subgroups, contributing to a more precise conceptualization of anomalies. In the context of the power generation plants dataset, anomalies can manifest through various factors, including malicious activities like fuel pilferage. 

The task of investigating anomaly is often referred to as either anomaly detection, novelty detection or outlier detection\cite{Pang}, and many techniques have been used for the purpose, either for specific domains such as finance \cite{anomaly_finance}, health \cite{samariya2023detection}, Internet of Things \cite{anomaly_cyber}, and so on,  or the generic ones \cite{anomaly_general, kumar_anomaly}. Furthermore, as the world witnesses the continuous influx of vast datasets, sophisticated machine learning, and deep learning techniques have found application in the field of anomaly detection. As an illustration, the power generation plant dataset employed in this study was previously employed in \cite{jecinta} to identify anomalies using supervised machine learning methods—namely, logistic regression, support vector machines, k-Nearest Neighbors, and the Multilayer Perceptron. The latter work showed that the Multilayer Perceptron outperformed other models. The same dataset was also used in \cite{atemkeng2023label} to investigate and detect anomalies using a sort of label-assisted auto-encoders with results outperforming the work in \cite{jecinta}. 
We investigated the same dataset with Generative Adversarial Networks (GANs),  GANs present multiple benefits for anomaly detection, primarily by generating high-quality and diverse samples that effectively capture the complexity and variability inherent in real data. GANs involve training a classifier that assigns a probability score to a sample, indicating whether it is categorized as "normal" or "anomalous". This differs from auto-encoders used in \cite{atemkeng2023label}, where the input is compressed into a latent space, and classification is made based on reconstruction errors using the test samples.

As indicated in \cite{guansong}, recent years have witnessed the remarkable proficiency of deep learning approaches, such as GANs, in learning complex representations in data, including high-dimensional, temporal, geographical, and graph data. This progress has significantly expanded the horizons of numerous learning tasks. Additionally, \cite{luo2022anomaly} has substantiated that GANs possess the capacity to learn the behavioural patterns inherent in typical data, given their ability to replicate intricate and high-dimensional data distributions. Consequently, this paper advocates the use of GANs for the purpose of anomaly detection within the power generation plants of the TeleInfra company, thereby complementing the research in \cite{jecinta, atemkeng2023label}.

This research employed a time series dataset that was obtained through the power consumption activities at TeleInfra Telecommunication company, being the case study. Consequently, it could be limited in respect of time constraints, since the data was collected in a specific time period. The rest of this work is organized as follows: Section \ref{s2} discusses the method used in the analysis; Section \ref{s3} presents the data and data augmentation strategy; Section \ref{s4} contains the results of the analysis, as well as the discussions; and Section \ref{s5} shows suggestions on the possible future direction of work in this line.

\section{Methods}
\label{s2}
\subsection{Generative Adversarial Network: Overview}
Anomaly detection methods have metamorphosed from machine learning methods into deep learning methods due to the continuous availability of datasets. GANs is a deep learning approach that represents an instance of modern generative modelling\textemdash it is typically used in replicating the distribution of datasets \cite{goodfellow}. GAN is based on game theory, compared to many other generative models which are based on optimization.  

The GANs algorithm comprises two key components: the "generator" and the "discriminator." The "discriminator" functions as a deep-learning classifier, trained to differentiate between real and artificially generated examples drawn from the input domain. On the other hand, the "generator," also a deep-learning model, takes a random latent vector from a normal distribution as input and aims to generate examples that convincingly deceive the discriminator into perceiving them as real. These two components learn in tandem, iteratively refining their capabilities until they reach a point of equilibrium where the generator consistently produces authentic-looking examples.

Also, after learning the true data distribution, GANs produce samples that are comparable to the training dataset. Thus, we train a classifier that gives a probability score of a sample as being \textit{normal} or \textit{anomalous}, rather than compressing the input into a latent space and classifying the test samples based on the reconstruction error.

\subsection{Building the GAN}
A noise sample $\mathbf{z}$ is provided as input to the generator neural network, $G$. Subsequently, the generator generates a synthetic sample, $G(\mathbf{z})$. Both the generated sample and the actual data $\mathbf{x}$ are then forwarded to the discriminator neural network,  $D$. The primary role of the discriminator is to produce a single value, representing the likelihood that the given input is a real sample. In this context, it outputs $D(\mathbf{x})$ when the initial input is real, and conversely, it outputs $D(G(\mathbf{z}))$ if the input is generated. The discriminator operates as a classifier. An adversarial framework is depicted in Figure \ref{fig1}.

\subsection{Mathematical Framework of the GANs}
The adversarial framework, as introduced by \cite{goodfellow}, is mathematically represented as a minimax optimization between the generator $G: \mathbb{R}^d \rightarrow \mathbb{R}^n$ and the discriminator $D: \mathbb{R}^n \rightarrow [0, 1]$, with a designated target loss function.
\begin{align}\label{eqn1}
    V(D, G) &= \mathbb{E}_{\mathbf{x} \sim p_{data}}[\log D(\mathbf{x})] + \mathbb{E}_{\mathbf{z} \sim p_{z}}[\log (1 - D(G(\mathbf{z}))].
\end{align}
The first term, $\mathbb{E}{\mathbf{x} \sim p{data}}[\log D(\mathbf{x})]$, represents the discriminator's assessment of real data. Conversely, the second term, $\mathbb{E}{\mathbf{z} \sim p{z}}[\log (1 - D(G(\mathbf{z}))]$, is the prediction of the discriminator's on a fake data.
\begin{figure}
\centering 
\includegraphics[width=0.5\textwidth]{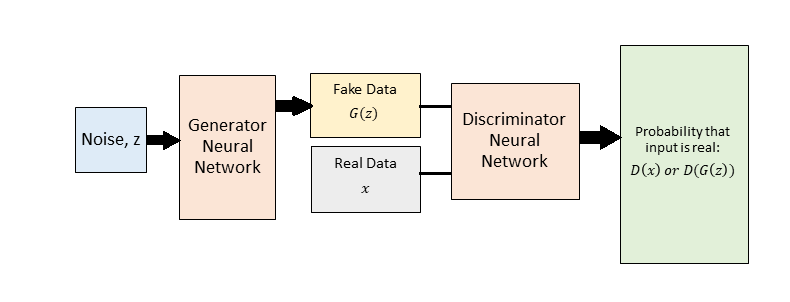}
\caption{Building Block of a GAN}
\label{fig1} 
\end{figure}
Bearing in mind Equation \eqref{eqn1} required  that we solve the minimax problem:
\begin{equation}
\begin{aligned}
\fontsize{5pt}{5pt}\selectfont
  \underset{G}{\min}\; \underset{D}{\max} \; V(D, G) &= \underset{G}{\min}\; \underset{D}{\max}\big (\mathbb{E}_{\mathbf{x} \sim p_{data}}[\log D(\mathbf{x})]  \\
  &+\mathbb{E}_{\mathbf{z} \sim p_{z}}[\log (1 - D(G(\mathbf{z}))] \big).\label{eqn2}
\end{aligned}
\end{equation}
The generator aims to maximize its probability of success, which involves causing the discriminator to make mistakes. Consequently, the generator seeks to minimize the value function defined by Equation \eqref{eqn2}. In contrast, the discriminator strives to minimize the generator's likelihood of success and, therefore, aims to maximize Equation \eqref{eqn2}. Specifically, the discriminator endeavors to minimize $D(G(\mathbf{z}))$ while maximizing $D(\mathbf{x})$. Algorithm \ref{alg:two} outlines the computational procedures employed in GAN.
\SetKwComment{Comment}{/* }{ */}
\begin{algorithm}
\caption{Algorithm of the GANs \cite{goodfellow}}\label{alg:two}
\KwData{Real data, $\mathbf{x}$ and hyperparamter, $k$}
\KwResult{Fake data, $\tilde{\mathbf{x}}$}
\For{each training iteration}{
\For{$k$ steps}{
  Sample $m$ noise samples $\{z_1, z_2, \dots, z_m \}$ and transform with Generator;\\
  Sample $m$ real samples $\{x_1, x_2, \dots, x_m \}$ from real data;\\
  Update the Discriminator by \textbf{ascending} the gradient:
$\big\Uparrow$ $\nabla_{\theta_d} \frac{1}{m}\displaystyle \sum_{i = 1}^m \Big[\log D\left(x^{(i)}\right) + \log \Big( 1 - D\left(G\left(z^{(i)}\right)\right) \Big)\Big] $;
  }
  Sample $m$ noise samples $\{z_1, z_2, \dots, z_m\}$ and transform with Generator;
Update the Generator by \textbf{descending} the gradient:
$\big\Downarrow$ $\nabla_{\theta_g} \frac{1}{m}\displaystyle \sum_{i = 1}^m \log \Big( 1 - D\left(G\left(z^{(i)}\right)\right) \Big) $;
}
\end{algorithm}
The gradient ascent of the loss function is employed to update the discriminator to maximize the cost, while the generator is updated in the opposite direction, using gradient descent, to minimize the cost. This aligns with the process depicted in Figure \ref{fig1}.

\subsection{Performance Evaluation Metrics}
To assess the performance of the model, four essential model evaluation metrics were used:
\begin{itemize}
\item  Confusion Matrix:
A matrix displaying the model's correct and incorrect predictions, comprising True Positive (TP), True Negative (TN), False Positive (FP), and False Negative (FN).
	\item Accuracy,  $Ac$: 
The probability of correct predictions is represented as the ratio of total correct predictions to the total number of observations in the test set. It is mathematically expressed as:
	\begin{align}\label{eqn3}
		Ac  = \frac{TP + TN }{TP+ FP+ TN+ FN}.
	\end{align}
	\item Sensitivity, $Sn$:
The ratio of correct positive predictions to the total number of positives. It is also referred to as Recall or True Positive Rate:
	\begin{align}\label{eqn4}
		Sn=\frac{TP}{TP+FN}.
	\end{align}
	\item Precision, $P$:
The ratio of correct positive predictions to the total number of positive predictions; also known as Positive Predictive Value:
	\begin{align}\label{eqn5}
		P=\frac{TP}{TP+FP}.
	\end{align}
	
	\item F$_1$-Score: It is defined by the harmonic mean of precision and recall:
	\begin{align}\label{eqn6}
		\text{$F_1$-Score}=2 \times \frac{Sn \times P}{Sn +P}.
	\end{align}

\end{itemize}

\section{Data Analysis}
\label{s3}
\subsection{Dataset}
The dataset for this study was sourced from TeleInfra Telecommunication Company's base stations in Cameroon, where generators served as the primary power source. It covers the period from September 2017 to August 2018, initially containing 6010 records. After data cleaning, it was reduced to 5905 records. The dataset includes variables such as "Running Time," "Power Type," "Consumption Rate," "GENERATOR CAPACITY (kVA)," and more. For a detailed dataset discussion, please refer to \cite{jecinta}.

\subsection{Pre-processing}
The dataset was painstakingly examined for possible defects like missing observations, and more importantly, relevant new variables were derived to corroborate the existing variables, and adequately represent some necessary measures in the bid to classify the labels into the anomaly and otherwise. As an example, the variable "Daily Consumption within a Period" was computed by dividing "Consumption HIS" by the "Number of Days." Similarly, "Running Time per Day" was determined by dividing "Running Time" by the "Number of Days," and "Daily Consumed Quantity between Visits" was calculated by dividing "Quantity Consumed between Visits" by the "Number of Days."

\subsection{Feature Importance}
Due to the numerous variables constituting the dataset, it becomes pertinent to rank the features based on their perceived contribution to the resulting model. 

\begin{figure}[h!]
\centering 
\includegraphics[width=0.53 \textwidth]{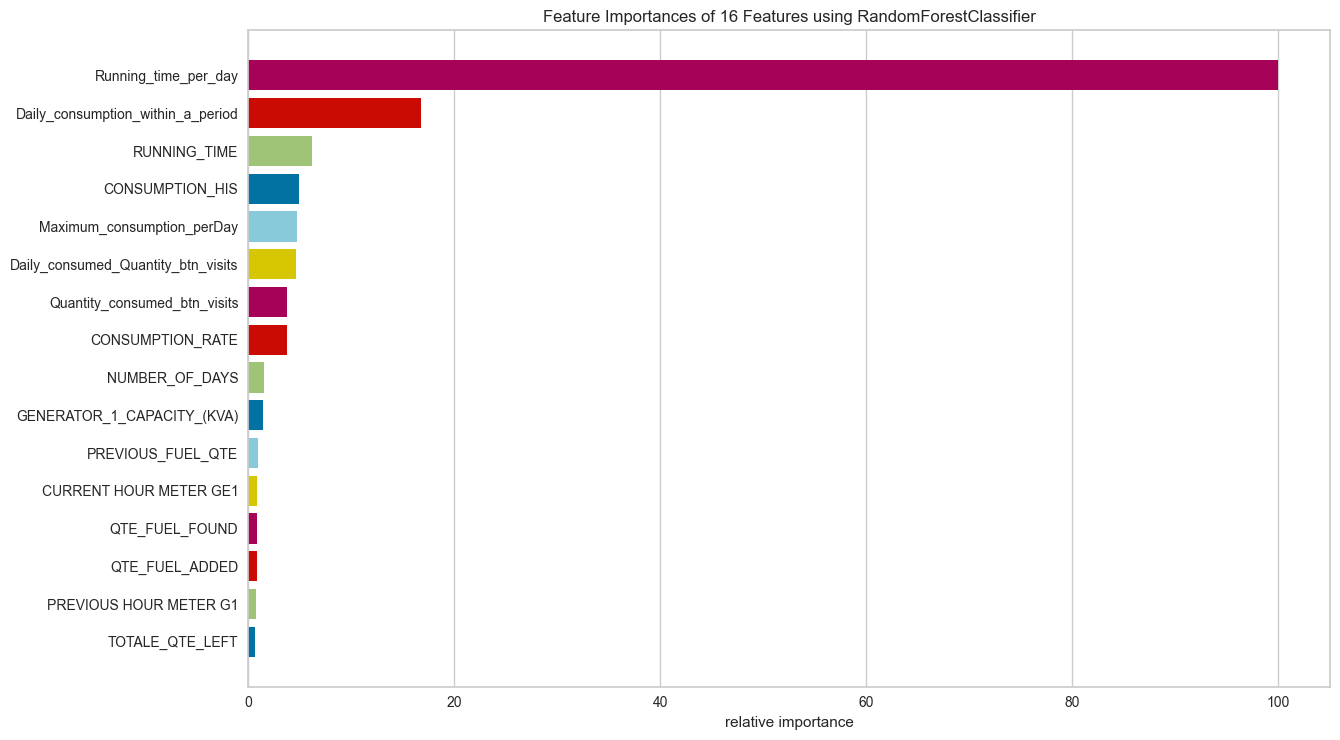}
\caption{Feature Importance}
\label{fig2} 
\end{figure}

In this task, the random forest classifier, as illustrated in Figure \ref{fig2}, was employed. Among all the features, "Running Time Per Day" exhibited the highest importance, followed by "Daily Consumption within a Period," "Running Time," "Consumption HIS," and "Maximum Consumption per Day," in descending order of significance. Additionally, it was observed that the feature with the least importance among all was "Total Quantity (QTE) Left."
\subsection{Anomaly Visualisation}
\begin{figure}[h!]
	\centering
	\begin{subfigure}{0.5\textwidth}
		\includegraphics[width=\textwidth]{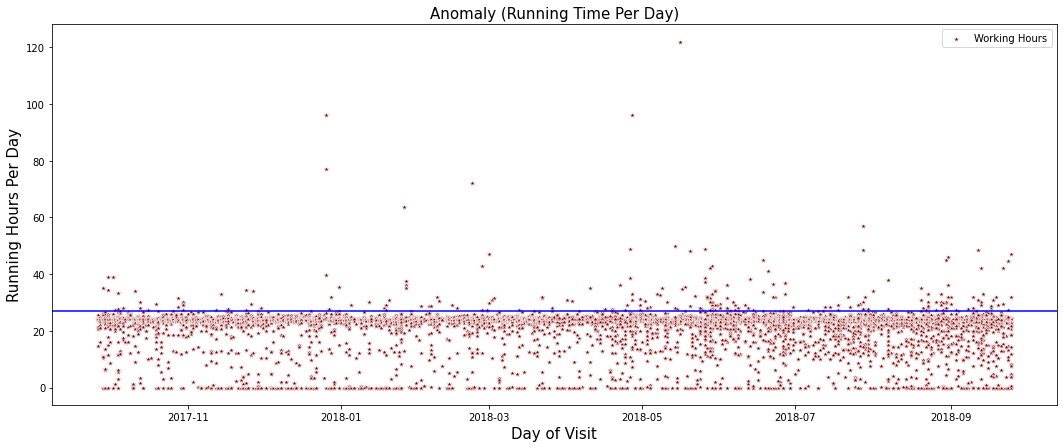}
		\caption{ Scatter plot of the Running Time Per Day}
\label{fig3b}
	\end{subfigure}
 	\begin{subfigure}{0.5\textwidth}
		\includegraphics[width=\textwidth]{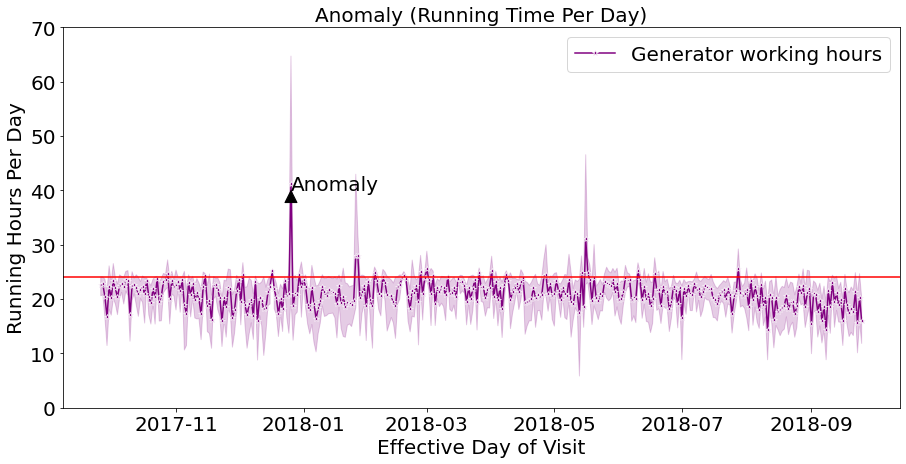}
		\caption{Time series plot  of the Running Time per Day}
		\label{fig3a}
	\end{subfigure}
	\caption{Anomaly visualisation from the Running Time Per Day at the Base Stations}
	\label{fig3}
\end{figure}

The scatterplot in Figure \ref{fig3b}, illustrates the distribution of daily running times for the power plants at TeleInfra telecommunication company's base stations. Notably, some data points surpass the 24-hour threshold in Figure \ref{fig3b}, identifying them as "anomalies." This suggests the presence of anomalies in fuel consumption concerning daily running times. Additionally, Figure \ref{fig3a} reveals that certain data points deviate significantly from the rest of the dataset. Both Figures \ref{fig3b} and \ref{fig3a}  are used to confirm the existence of an anomaly in the dataset.

\subsection{Label Classification}
Following feature selection, the dataset's target variable was determined by categorizing data points into either anomaly or normal cases based on variables like "Running Time per Day" and "Maximum Consumption Per Day," among others. There are 3829 records (64.88\%) classified as normal and 2073 (35.12\%) as anomalies. Additionally, this highlights the presence of a class imbalance in the data labels.

\subsection{Correlation Analysis}
\begin{figure}[h!]
\centering 
\includegraphics[width=0.55\textwidth]{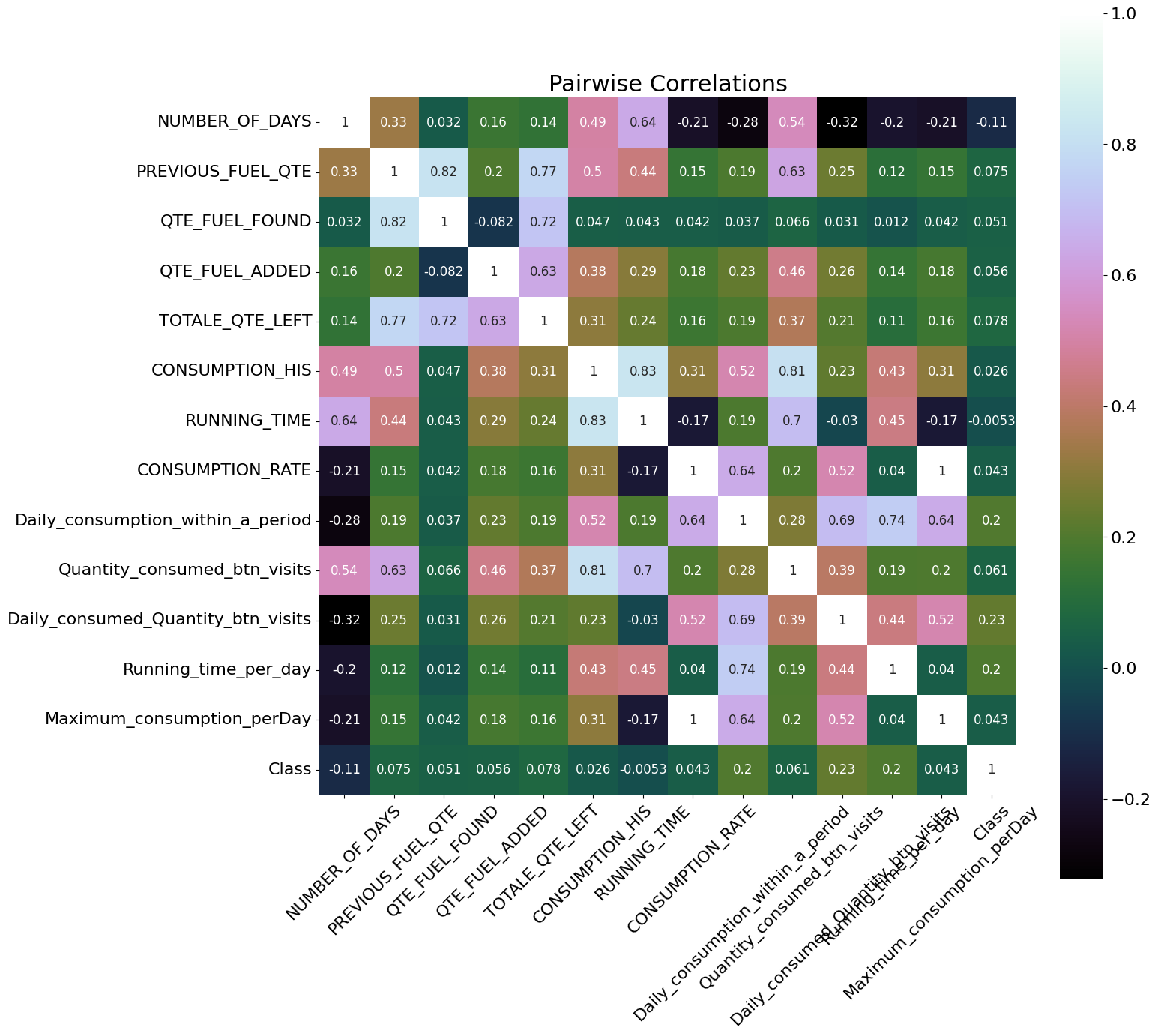}
\caption{Correlation Matrix }
\label{fig5} 
\end{figure}

 Examining the correlation matrix (Figure \ref{fig5}) reveals notable correlations, such as the strong positive correlation of 0.83 between "Consumption HIS" and "Running Time." This suggests that higher Running Time corresponds to increased consumption. Consequently, these variables are deemed relevant for analyzing fuel consumption patterns and detecting anomalies at the base stations.


\subsection{Data Augmentation}
As a deep learning model, GAN is data-demanding;  its performance  is dependent on the availability of a large volume of data. Consequently, a deep learning model requires consistent, accurate, and complete data \cite{data_challenge_Munappy2019DataMC}, making it inevitable to develop techniques for increasing existing datasets, in the scenarios of limited datasets. One of the viable methods that is used in the literature is data augmentation, although it is commonly used for images and text data. 

Recent research used different augmentation techniques such as auto-encoder \cite{autoencode_data_augmentation} and mask Token Replacement (MTR) \cite{rethinking_data_augmentation} for tabular data. In this study, the dataset, which had been reduced to 5902 observations after data cleaning, was augmented by adding a random value from a uniform distribution (ranging from 0 to the standard deviation of each feature) to the existing values in each feature's row. As a result, the augmented dataset now comprises a total of 187281 records, preserving the original features.

\section{Results and Discussion}
\label{s4}
As recommended in previous research on anomaly detection \cite{anomaly_detection_network}, we followed specific strategies for optimizing our generator and discriminator networks. To be precise, we used the Adam optimizer for the generator network while opting for the stochastic gradient descent optimizer for the discriminator network. The generator network was designed, featuring a total of five dense layers. These layers were equipped with the $\tanh$ activation function, and the binary cross-entropy loss function was chosen for its loss calculation.  Conversely, the discriminator network exhibited a slightly different structure. It comprised six dense layers, with a crucial addition of dropout layers. These dropout layers played a pivotal role in ensuring that the model did not excessively adapt to the training data. Furthermore, we employed the sigmoid activation function in the final layer of the discriminator network to facilitate its discriminative capabilities.

\begin{figure}[h!]
	\centering
	\begin{subfigure}{0.5\textwidth}
		\includegraphics[width=\textwidth]{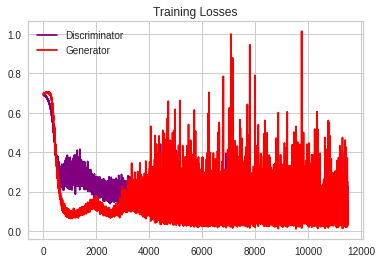}
		\caption{Training Losses without data augmentation}
		\label{fig6a}
	\end{subfigure}
	\begin{subfigure}{0.5\textwidth}
		\includegraphics[width=\textwidth]{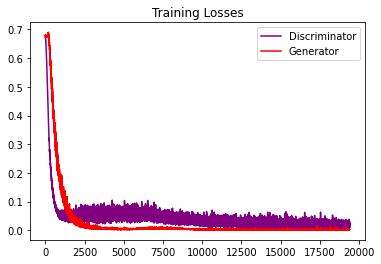}
		\caption{Training Losses with data augmentation}
\label{fig6b}
	\end{subfigure}
	\caption{The training losses of the GANs for anomaly detection with and without data augmentation.}
	\label{fig6}
\end{figure}

 The training losses in Figure \ref{fig6}  show how both the discriminator and generator models converge. In Figure \ref{fig6a}, we can observe the training loss of the GAN model trained on the dataset without data augmentation. A notable observation is that the training errors for both the discriminator and the Generator do not exhibit stabilization; instead, they fail to converge asymptotically toward zero as training progresses. This is due to the constrained quantity of data employed for training the model. The limited dataset size contributes to the challenges in achieving the desired convergence behaviour. Figure \ref{fig6b} provides insight into the training dynamics, displaying a noteworthy contrast. The training loss for both the discriminator and generator exhibits a consistent and smooth convergence toward zero. This favourable convergence behaviour can be attributed to the GAN being trained on an augmented dataset. The augmentation of the dataset plays a crucial role in promoting convergence and, consequently, achieving a level of performance that meets acceptable standards. By leveraging the augmented dataset, the GAN is better equipped to navigate the training process, leading to improved stability and convergence in the training loss.

\begin{table}[!h]
    \centering
     \caption{The evaluation metrics for the GAN model indicate that the augmented data version outperforms the non-augmented counterpart.}
    \begin{tabular}{c|c|c}
    \hline
    \hline
        Metrics & Without data augmentation & With data augmentation\\
        \hline
        \hline
        Accuracy Score & 0.6645 & 0.9899\\
       Precision & 0.5455 &  0.78538\\
       Recall & 0.0151 & 0.7966\\
       $F_1$ Score & 0.0294 & 0.6457\\
       \hline
    \end{tabular}
    \label{tab2}
\end{table}

\begin{figure}
	\centering
		\includegraphics[width=0.5\textwidth]{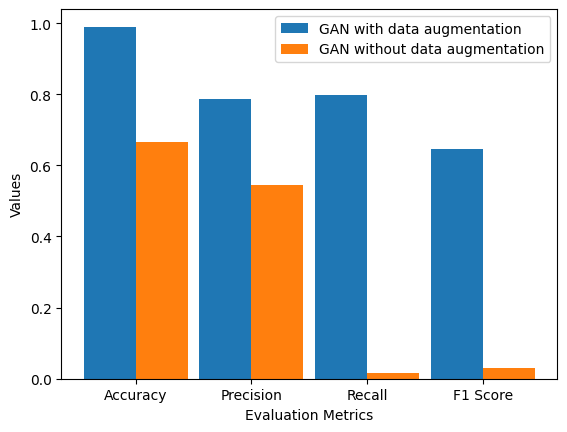}
	\caption{The evaluation metrics for the GAN model indicate that the augmented data version outperforms the non-augmented counterpart.}
	\label{figxx}
\end{figure}
Figure \ref{figxx} and Table \ref{tab2} provide a comprehensive comparison of the performances between the GAN models with and without data augmentation. The comparison highlights a significant disparity in performance outcomes. Specifically, the GAN model that incorporates data augmentation demonstrates superior performance across all the evaluation metrics employed in this study. This observation underlines the effectiveness of data augmentation in enhancing the capabilities of deep learning models.
\section{Conclusion}
\label{s5}
Due to the inconsistent power supply in Cameroon, TeleInfra, a telecommunications company, has turned to alternative power sources, predominantly relying on generators. While generating power using generators, the company noticed irregularities in fuel consumption patterns. These anomalies were identified by assessing variables such as daily running time, daily consumption rate, and similar metrics.

The dataset was collected on the different features relevant to detecting anomalies in the power generation plant. Building upon the groundwork in \cite{jecinta}, who employed various machine learning techniques including K-Nearest Neighbors, Logistic Regression, Multilayer Perceptron, and Support Vector Machines, and further extending the research conducted in \cite{atemkeng2023label}, which introduced a label-assisted auto-encoder approach, our study investigated  GANs for anomaly detection. GAN trains a  classifier that allocates a probability score to each sample, thereby determining its classification as either "normal" or "anomalous". This contrasts with the approach adopted in  \cite{atemkeng2023label}, where the input data is encoded into a latent space, and the classification of test samples is predicated on the evaluation of reconstruction errors.
During the GAN training phase, an initial observation revealed suboptimal performance, potentially attributed to the limited size of the dataset. Consequently, to address this concern, data augmentation was implemented on the initial dataset to generate additional data points. Subsequently, when employing this augmented dataset, the GAN demonstrated superior performance compared to the models introduced in the works in \cite{jecinta, atemkeng2023label} in terms of accuracy. However, it's noteworthy that in certain metrics, such as precision, the GAN's performance did not match its performance in accuracy.
The evaluation metrics effectively substantiate the advantage gained by incorporating augmented data during the model's training process. The augmented data's influence is evident in the augmented model's ability to consistently achieve superior results, thereby affirming the value of data augmentation as a strategy for optimizing the GAN's performance in various aspects of anomaly detection.




\section*{Acknowledgment}
The authors would like to express their gratitude to the reviewers for their valuable comments. Additionally, MA extends his appreciation to Rhodes University for its support in conducting this research.
\bibliographystyle{IEEEtran}
\bibliography{the_references}


\end{document}